\documentclass[12pt]{article}

\usepackage{sbc-template}

\usepackage{graphicx,url}

\usepackage[utf8]{inputenc}  
\usepackage{float}
     
\title{Application of Attention Mechanism with Bidirectional Long Short-Term Memory (BiLSTM) and CNN for Human Conflict Detection using Computer Vision.}

\author{Erick da Silva Farias\inst{1}, Eduardo Palhares Júnior \inst{2}}

\address{Instituto Federal de Educação, Ciência e Tecnologia do Amazonas (IFAM) \\ \textit{Campus} Manaus Zona Leste --
 Manaus, AM -- Brasil
  \email {edsfrlinux@gmail.com, eduardo.palharesjr@ifam.edu.br}
}

\begin{document} 

\maketitle

\begin{abstract}
The automatic detection of human conflicts through videos is a crucial area in computer vision, with significant applications in monitoring and public safety policies. However, the scarcity of public datasets and the complexity of human interactions make this task challenging. This study investigates the integration of advanced deep learning techniques, including Attention Mechanism, Convolutional Neural Networks (CNNs), and Bidirectional Long Short-Term Memory (BiLSTM), to improve the detection of violent behaviors in videos. The research explores how the use of the attention mechanism can help focus on the most relevant parts of the video, enhancing the accuracy and robustness of the model. The experiments indicate that the combination of CNNs with BiLSTM and the attention mechanism provides a promising solution for conflict monitoring, offering insights into the effectiveness of different strategies. This work opens new possibilities for the development of automated surveillance systems that can operate more efficiently in real-time detection of violent events.
\end{abstract}

\section{Introduction}


Violence is a complex phenomenon that permeates the history of humanity, manifesting itself in different ways and in different contexts. Since the beginning of civilization, violence has been present in wars, territorial conflicts, and power disputes. Over time, new manifestations emerged, such as domestic violence, urban crime, and terrorist attacks. Violence manifests itself in seemingly trivial situations, such as fights in bars or traffic conflicts. These episodes reflect social tensions, accumulated frustrations, and, often, the lack of adequate conflict resolution mechanisms. The culture of aggression and the normalization of violence in social relationships can intensify these situations, creating a cycle that is difficult to break.


In many contexts, social inequality, poverty, and marginalization also fuel violence, creating an environment conducive to organized crime and urban violence. Thus, violence in today's world is a multifaceted phenomenon that requires a critical and multidisciplinary approach to understand it and, above all, combat it. Analysis of its historical, social, and cultural roots is fundamental to developing effective prevention and intervention strategies.


Surveillance cameras are widely used in commercial establishments, homes, industries, schools, and public places. These cameras are intended to assist agents who monitor the location, however, this type of conventional monitoring is not very effective when hundreds of cameras are deployed because of human involvement, because identifying incidents using conventional cameras becomes an inefficient task.


An efficient way to identify incidents via a surveillance camera would be through computer vision, because images from the CCTV system can be linked to a trained deep learning model to make inferences about incidents related to violence between humans in real time. This approach to using computer vision is relevant as it will eliminate the cost of surveillance by humans. But for this to work, it is necessary to carry out tests, collect images to train the model, compare deep learning models, and other adjustment processes to refine the human conflict detection system.


With respect to data collection, it is important that the data set has a significant volume, with variance in class data and good resolution. According to \cite{Dashdamirov2024}, for effective algorithm training, the collection and labeling of a vast volume of data is essential. Although there are public sets of videos available, there is still a significant need to expand the amount of this data. Furthermore, aspects such as video resolution, frame frequency, lighting conditions, and camera angles vary greatly. These differences complicate the development of models that are both robust and capable of generalizing appropriately.

The use of Deep Learning in the context of human conflict monitoring is relatively new, because the data available publicly has a small volume and has low quality in the video frames.
\cite{Dashdamirov2024} evaluates deep learning techniques in detecting violence in videos, highlighting that increasing the dataset from 500 to 1,600 videos improves the average accuracy of the models by 6\%. It demonstrates the importance of large data sets and transfer learning for more effective surveillance systems.

\cite{Datta2002} analyzed the trajectory of movements and orientation of body limbs to detect violent behavior. \cite{Nguyen2005} introduced a hierarchical hidden Markov model (HHMM), showing that it can be useful for recognizing aggressive attitudes, especially through a standard HHMM approach aimed at identifying violence.

\cite{kim2009} combined probabilistic Principal Component Analysis (PCA), used to identify flow patterns in local areas, with Markov Random Fields (MRF), which help maintain global model coherence. On the other hand, \cite{Mahadevan2010} argued that optical flow-based representations are not suitable for detecting unusual changes in both appearance and motion. They proposed a technique that identifies violent scenes by evaluating elements such as the presence of blood, flames, intensity of movement and sound volume.


\section{Methodology}


In this chapter the methodology will be presented. In \ref{sec2_1}, computer vision was discussed. In section \ref{sec2_2} Deep Learning and Neural Networks were covered, LSTM and BiLSTM in the subtopics and in session \ref{sec2_3} about the Attention Mechanism.

\subsection{Computer Vision}\label{sec2_1}



Computer vision is an area of artificial intelligence (AI) that deals with developing methods that allow computers to acquire, process and interpret visual information from the real world, with the aim of making decisions or providing recommendations \cite{Szeliski2010}.


The main challenges of computer vision in videos involve the need to identify and classify objects and actions in dynamic environments, such as recognizing human behavior patterns or detecting specific events, such as conflicts, aggressions or complex interactions.

\subsubsection{Neural Network Models in Computer Vision}

Convolutional neural network (CNN) models are widely used in computer vision due to their ability to extract hierarchical spatial features from images and videos. CNNs operate by applying filters (or convolutions) to the image to extract local features such as edges, textures and shapes. These models are efficient for tasks such as object detection, scene recognition and action identification in videos \cite{LeCun2015}.


For video analysis, CNNs are often combined with temporal models such as Long Short-Term Memory (LSTM) networks in order to capture dynamic information over time, effectively integrating spatial and temporal learning \cite{Simonyan2014}.

\subsection{Deep Learning and Neural Networks}\label{sec2_2}

Deep learning is a subfield of artificial intelligence that relies on deep neural networks to perform complex recognition, classification, and prediction tasks. These networks are composed of multiple layers of processing, allowing them to learn hierarchical representations of data such as images, text and temporal sequences.

\subsubsection{Convolutional Neural Networks (CNNs)}

\begin{figure}[ht]
\centering
\includegraphics[width=1\textwidth]{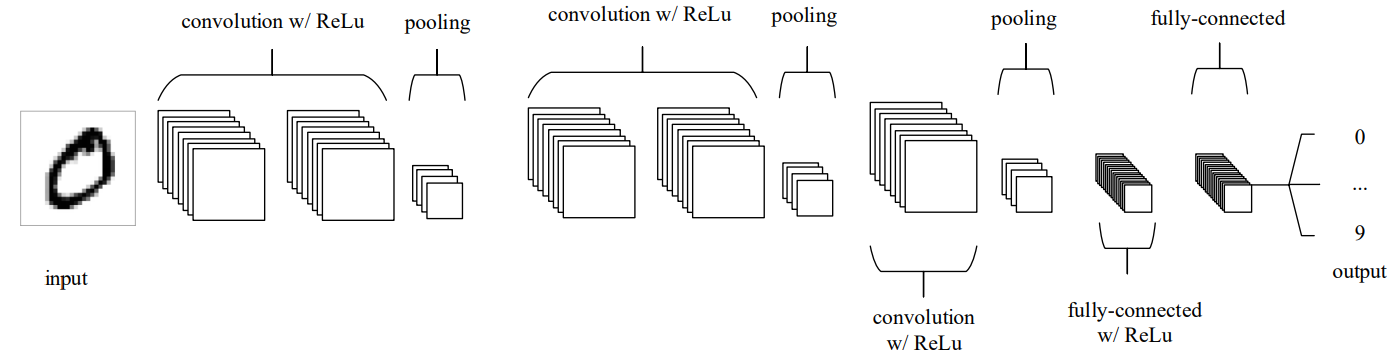}
\caption{An example of CNN architecture \cite{O’Shea2015}}
\label{fig:exampleFig1}
\end{figure}

Convolutional neural networks (CNNs) are a class of deep neural networks that have been widely used in computer vision tasks due to their ability to learn efficient representations of visual data. Figure \ref{fig:exampleFig1} shows a diagram that represents the architecture of a CNN. They are composed of convolutional layers, pooling layers, and fully connected layers. CNNs are effective in extracting spatial features from images, which allows them to detect patterns, such as edges, textures and shapes \cite{LeCun2015}. 

When applied to video analysis, CNNs can be used to detect moving objects, such as humans, vehicles, or any other type of interest. These networks are also capable of detecting complex behaviors and interactions by extracting spatial features from each video frame and learning sequential representations.

\subsubsection{Long Short-Term Memory (LSTM)}

\begin{figure}[ht]
\centering
\includegraphics[width=1.0\textwidth]{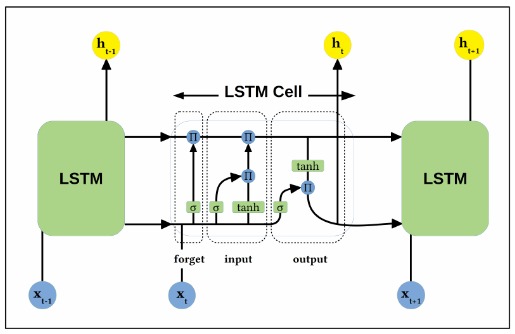}
\caption{Basic Long-Short Term Memory (LSTM) architecture \cite{Shenfield2020}}
\label{fig:exampleFig2}
\end{figure}

Long Short-Term Memory (LSTM) is a recurrent neural network (RNN) architecture designed to model temporal dependencies in sequential data. Figure \ref{fig:exampleFig2} shows the LSTM architecture diagram. LSTMs have memory cells that allow the retention of information over time, overcoming the problem of gradient fading that limits other traditional RNNs \cite{Hochreiter1997}.

In the context of videos, LSTMs are used to model temporal sequences, capturing long-term dependencies between frames or between events that occur at different times, which is essential for detecting human behaviors, such as identifying actions or interactions.

\subsubsection{Bidirectional LSTM (BiLSTM)}

\begin{figure}[ht]
\centering
\includegraphics[width=.9\textwidth]{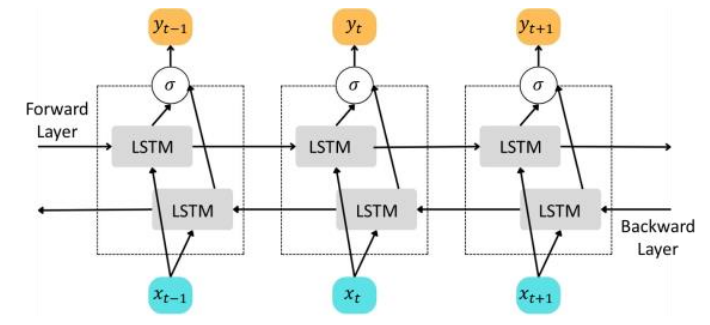}
\caption{Architecture of Bi-LSTM network \cite{Raihan2023}}
\label{fig:exampleFig3}
\end{figure}

Bidirectional LSTM (BiLSTM) is a variation of LSTMs that processes the sequence of inputs in both the forward and reverse directions, which allows the model to have a more complete understanding of temporal dependencies. Figure \ref{fig:exampleFig3} shows the BiLSTM architecture diagram.

 \newpage
 
This is particularly useful in videos, where context can be influenced by both past and future events. BiLSTM can be applied to detect conflicts or dynamic events, allowing the model to understand not only what happens before, but also what happens after a certain event, improving accuracy in identifying complex patterns \cite{Graves2013}.

\subsection{Attention Mechanism}\label{sec2_3}

The Attention Engine is a fundamental technique in the field of deep learning, used to improve the ability of models to focus on the most relevant parts of input during processing. Instead of treating all elements of the input equally, the Attention Engine allows the model to learn to allocate greater weight to the most informative parts of the input, improving the efficiency and accuracy of predictions. This technique was initially proposed by \cite{Bahdanau2015}, and later refined into models such as the Transformer proposed by \cite{Vaswani2017}, which use attention mechanisms as their central basis.

\subsubsection{Types of Attention Mechanisms}

There are different types of attention mechanisms, each with its own characteristics and forms of implementation. Below, we discuss two main types of attention mechanisms: the mechanism of attention with weighted sum and the multi-head attention mechanism.
The attention with weighted sum focuses on assigning different weights to input features, allowing the model to prioritize more relevant information while processing sequences. On the other hand, the multi-head attention mechanism improves the model's ability to focus on various parts of the input simultaneously by using multiple attention heads. However, for the purposes of this article, the discussion will be limited to the attention mechanism with weighted sum.


The mechanism we use in our model is a simple version of attention, often called weighted sum attention. In this type of mechanism, the model calculates an attention score for each element of the input sequence, using a dot product between the input vector $x_i$ and a weight vector \textit{W} learned during training. The attention score $e_i$ for each element $X_i$ is given by the formula:

\[ e_i = \tanh(W^T x_i + b) \]


\noindent where:


\begin{itemize}
    \item $e_i$ is the attention score of element $x_i$
    \item \textit{W} is the weight vector.
    \item $b$ is the bias.
    \item  $tanh$ is the activation function that helps limit the value of the attention score.
\end{itemize}


Next, the $e_i$ attention score is normalized using the \textbf{softmax} function, generating the weights of $a_i$:

\[ a_i = \frac{e_i}{\sum_je_j} \]


\noindent Where $a_i$ are normalized \textbf{attention weights} that indicate the relevance of each element of the input. These weights are applied to the input $x_i$ generating a weighted sum of the inputs, as shown in the formula:

\[ output = \sum_i  a_ix_i\]


This Simple Attention Mechanism allows the model to focus on the most relevant parts of the input, assigning greater weights to inputs that are more informative for the task.

\begin{figure}[ht]
\centering
\includegraphics[width=.4\textwidth]{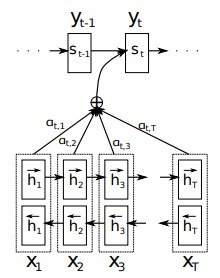}
\caption{Illustrative image of attention mechanism weighting \cite{Bahdanau2015}}
\label{fig:exampleFig4}
\end{figure}

This type of attention is useful in many image sequence processing tasks, where it is necessary to highlight specific regions or moments of the input \cite{Bahdanau2015}.

\subsubsection{Designed Architecture for Conflict Detection}


The model developed for classifying image sequences is based on a deep neural network designed to capture both spatial and temporal features from the data. The input consists of a sequence of 15 images, each with a size of 100x100 pixels and 3 color channels (RGB), initially processed by a convolutional layer (CNN). The first processing step applies a TimeDistributed layer, allowing the convolutional network to treat each image independently within the sequence.

To prevent overfitting, the model uses a Dropout layer immediately after this step, helping to improve generalization. Next, the architecture includes a Bidirectional LSTM layer, enabling the model to consider both past and future contexts of the image sequence, better capturing the temporal relationships between frames.

An Attention layer is applied afterward, allowing the model to perform weighting of the most relevant images within the sequence, giving more importance to certain frames to improve the analysis accuracy. The outputs of this layer are passed through several Dense layers, each followed by a new Dropout layer to ensure regularization of the model. Finally, the model includes a dense layer with two neurons, responsible for binary classification.

\begin{figure}[ht]
\centering
\includegraphics[width=1\textwidth]{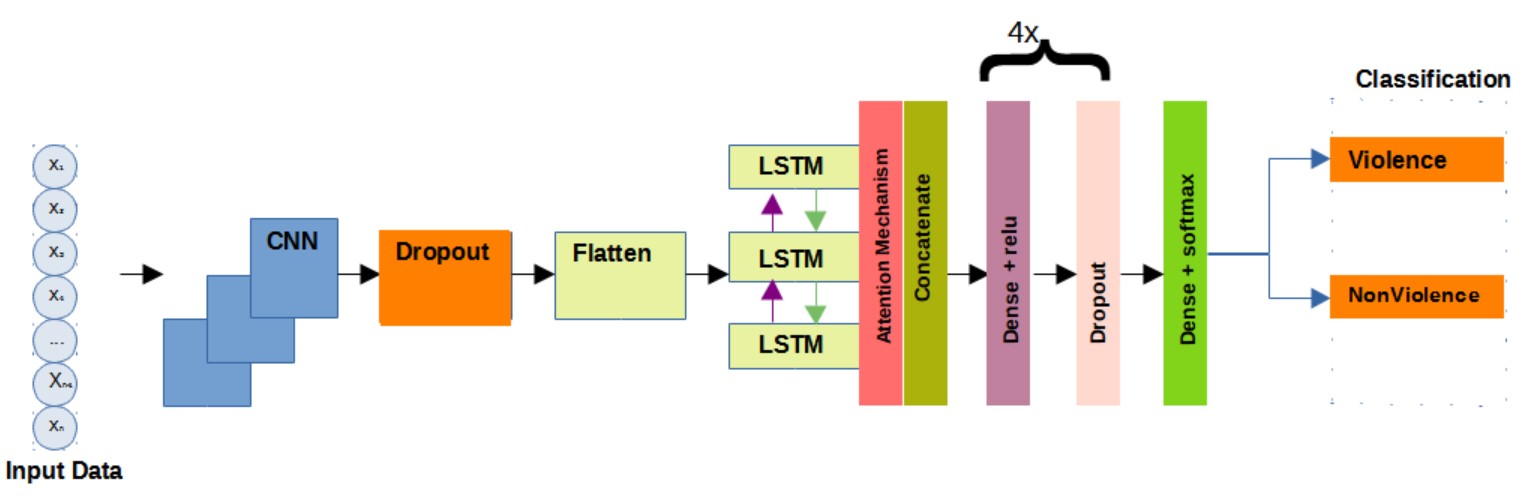}
\caption{Architecture implemented for the experiments. }
\label{fig:exampleFig5}
\end{figure}

Figure \ref{fig:exampleFig5} shows the diagram of the model architecture, illustrating the layers and the data flow throughout the process.

This combination of convolutional, recurrent, and attention layers allows the model to extract and learn complex, dynamic information from the images and their temporal sequences, providing a robust approach for the classification task.

\section{Results and Discussion}


The experiments were performed with the \textit{MobileNetV2}, \textit{DenseNet121} and \textit{InceptionV3}. It was also used \textit{BiLSTM} in conjunction with the models. The experiments were performed with and without attention mechanism. Regarding the parameters, there were variations in the minimum learning rate (\textit{min\_lr}) and batch size (\textit{batch\_size}). 

\begin{table}[ht]
\centering
\begin{tabular}{|c|c|c|c|c|c|}
\hline
\textbf{ID} & \textbf{Model} & \textbf{Attention} & \textbf{min\_lr} & \textbf{batch\_size} & \textbf{Accuracy (\%)} \\ \hline
1 & MobileNetV2 & No & 0.0005 & 128& 94.25 \\ \hline
2 & DenseNet121 & No & 0.0005 & 128& 94.75 \\ \hline
3 & InceptionV3 & No & 0.0005 & 128& 94.25\\ \hline
4 & MobileNetV2 & No & 0.00005 & 64& 89.00 \\ \hline
5 & DenseNet121 & No & 0.00005 & 64& 93.75 \\ \hline
6 & InceptionV3 & No & 0.00005 & 64& 91.00 \\ \hline
7 & MobileNetV2 & Yes & 0.0005 & 128& 93.25 \\  \hline
8 & DenseNet121 & Yes & 0.0005 & 128& 92.50 \\ \hline
9 & InceptionV3 & Yes & 0.0005 & 128 & 91.75 \\ \hline
10 & MobileNetV2 & Yes & 0.00005 & 64& 96.50 \\ \hline
11 & DenseNet121 & Yes & 0.00005 & 64& 95.50\\ \hline
12 & InceptionV3 & Yes & 0.00005 & 64& 94.25 \\ \hline
\end{tabular}
\caption{Training performed details with models \textit{MobileNetV2}, \textit{DenseNet121} and \textit{InceptionV3}.}
\label{tabela:experimentos}
\end{table}


Table \ref{tabela:experimentos} presents the details of 12 experiments performed with the models \textit{MobileNetV2}, \textit{DenseNet121} e \textit{InceptionV3}. In experiments, the variable of interest was the accuracy obtained during training, which varied according to the use of the attention mechanism, the minimum learning rate and the size of the lot.


In relation to the experiments without \textit{Attention Mechanism}, in the experiment with the minimum learning rate of 0.0005 and batch size 128, the \textit{DenseNet121} model obtained the highest accuracy, with 94.75\%, followed by \textit{ MobileNetV2} with 94.25\% and \textit{InceptionV3} with 94.25\%. When the minimum learning rate was reduced to 0.00005 and the batch size was adjusted to 64, \textit{MobileNetV2} had the lowest accuracy among all experiments at 89.00\%, while \textit{DenseNet121} had a slight drop in the value of accuracy, with 93.75\% and \textit{InceptionV3} had an accuracy of 91.00\%. Using \textit{Attention Mechanism}, the models showed a slight drop in accuracy with the minimum learning rate of 0.0005 and lot size 128. The accuracy of \textit{DenseNet121} was 93.25\%, \textit{DenseNet121} It was 92.50\%, and that of \textit{InceptionV3} was 91.75\%.

\begin{table}[ht]
\centering
\begin{tabular}{|c|c|c|c|}
\hline
 \textbf{Model} & \textbf{Accuracy (\%)} & \textbf{F1-Score (Class 0)} & \textbf{F1-Score (Class 1)}  \\ \hline
MobileNetV2  & 96.50 & 96.00 & 97.00  \\ \hline
DenseNet121  & 95.50 & 95.00  & 96.00 \\ \hline
InceptionV3 & 94.25 & 94.00 & 94.00 \\ \hline

\end{tabular}
\caption{Best results of model performance metrics \textit{MobileNetV2 }, \textit{DenseNet121 } and \textit{InceptionV3 }.}
\label{tabela:melhores_experimentos}
\end{table}


When the minimum learning rate was reduced to 0.00005 and the batch size was adjusted to 64, the performance was superior compared to the no-attention experiments.
The \textit{MobileNetV2} model achieved the best accuracy, with 96.50\%, followed by \textit{DenseNet121} with 95.50\%, and \textit{InceptionV3} with 94.25\%. In Table \ref{tabela:melhores_experimentos}, the best accuracies of each model are presented, and all models have good results with the accuracy and \textit{F1-Score} performance metrics.

\begin{figure}[H]
\centering
\includegraphics[width=.6\textwidth]{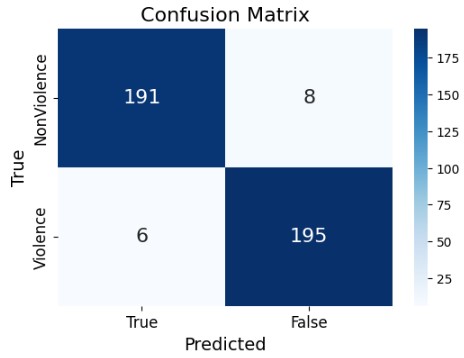}
\caption{Best performance confusion matrix}
\label{fig:matriz_conf}
\end{figure}


According to the confusion matrix of experiment 10, shown in the figure \ref{fig:matriz_conf}, it is observed in the axis of prediction that the nonviolence and violence classes hit almost all tests.

\begin{figure}[H]
\centering
\includegraphics[width=1.0\textwidth]{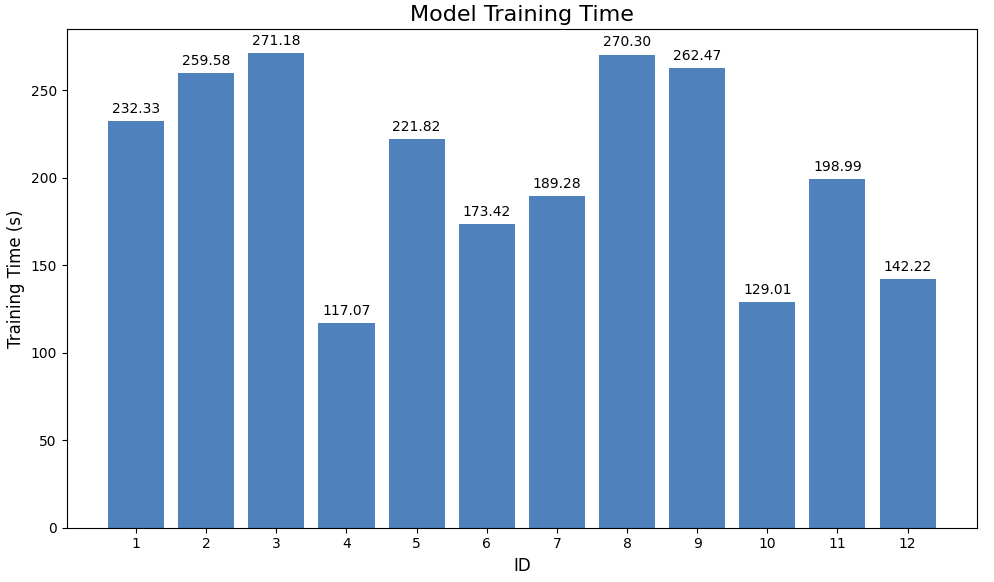}
\caption{Training time of all experiments.}
\label{fig:tempo}
\end{figure}


According to the figure \ref{fig:tempo}, and the reference of the experiments in the table \ref{tabela:experimentos}, it is observed that \textit{Attention Mechanism} did not extended training time. For example, experiments 1,2,3,7,8 and 9 have the same parameter settings. Experiments ID, 1,2 and 3 do not have \textit{Attention Mechanism} and experiments 7,8 and 9 have \textit{Attention Mechanism}. In other words, the mechanism has no influence on training time. Another relevant analysis that can be observed in Figure \ref{fig:tempo} is that training time decreases in experiments with reduced batch sizes.

\begin{figure}[H]
    \centering
    \begin{minipage}{0.45\textwidth}
        \centering
        \includegraphics[width=\linewidth]{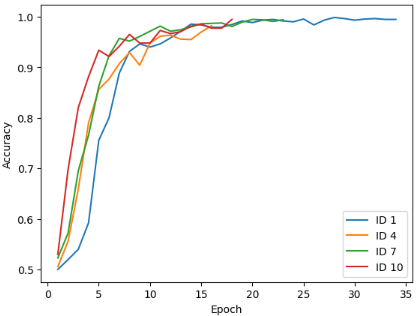}
        \vspace{0.1cm} 
        {\small \textnormal{(a) Accuracy in relation to the epochs of all training of the model \textit{MobileNetV2}}} 
    \end{minipage}%
    \hfill
    \begin{minipage}{0.45\textwidth}
        \centering
        \includegraphics[width=\linewidth]{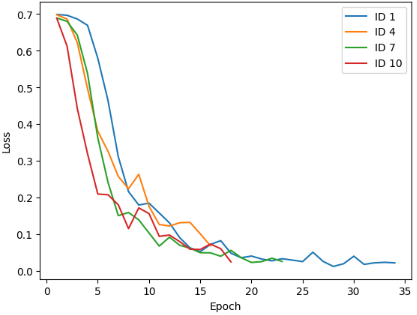}
        \vspace{0.1cm} 
        {\small \textnormal{(b) Error in relation to the epochs of all training of the model \textit{MobileNetV2}}} 
    \end{minipage}
    
    \vspace{0.3cm} 
    
    \begin{minipage}{0.45\textwidth}
        \centering
        \includegraphics[width=\linewidth]{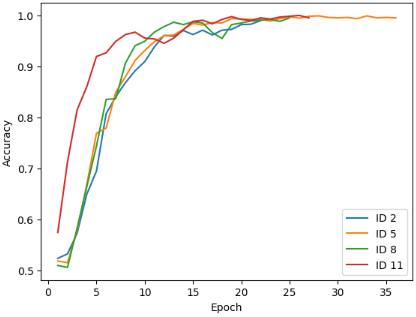}
        \vspace{0.1cm} 
        {\small \textnormal{(c) Accuracy in relation to the epochs of all training of the model \textit{DenseNet121}}} 
    \end{minipage}%
    \hfill
    \begin{minipage}{0.45\textwidth}
        \centering
        \includegraphics[width=\linewidth]{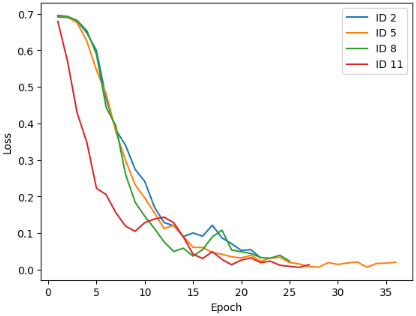}
        \vspace{0.1cm} 
        {\small \textnormal{(d) Error in relation to the epochs of all training of the model \textit{DenseNet121}}} 
    \end{minipage}

    \vspace{0.3cm} 

     \begin{minipage}{0.45\textwidth}
        \centering
        \includegraphics[width=\linewidth]{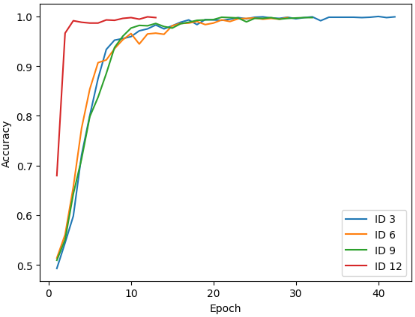}
        \vspace{0.1cm} 
        {\small \textnormal{(e) Accuracy in relation to the epochs of all training of the model \textit{InceptionV3}}} 
    \end{minipage}%
    \hfill
    \begin{minipage}{0.45\textwidth}
        \centering
        \includegraphics[width=\linewidth]{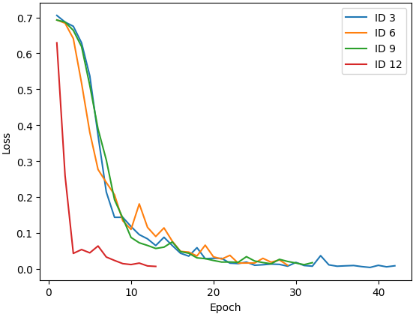}
        \vspace{0.1cm} 
        {\small \textnormal{(f) Error in relation to the epochs of all training of the model \textit{InceptionV3}}} 
    \end{minipage}
    
    \caption{Training of all experiments performed}
    \label{fig:matrix}
\end{figure}


In relation to the accuracy graph in Figure \ref{fig:matrix}, knowing that experiment 4 does not use the mechanism, it completes training faster, in relation to the other experiments. Another relevant point in the graph of Figure \ref{fig:matrix}, is that experiment 1 also does not use the mechanism and delay to finish the training. This indicates that the application \textit{Attention Mechanism} has no influence on model training time.


It can be observed that in all the error charts in the figure \ref{fig:matrix}, the reduced batch experiments with the mechanism have found the best minimum. These experiments sought the fastest minimums in the first times, as most of the batches gradients went to a specific direction. Another pattern of experiments 10, 11 and 12. It was that, due to the reduction of the steering speed to the minimum, the randomness led to directions in which the error increased slightly to later find better minimums.

\section{Conclusion and Future Works}
This study presented a detailed analysis of the application of Deep Learning models for violence detection in videos, focusing on the comparison of three popular architectures: MobileNetV2, DenseNet121, and InceptionV3. The results obtained demonstrated that all models were effective, with MobileNetV2 standing out by achieving the highest accuracy of 96.50\%, especially when the batch size was reduced and the attention mechanism was applied.

Through the experiments conducted, it was observed that the selection of appropriate parameters, such as learning rate and batch size, is crucial for optimizing the model's performance. Additionally, although the Attention Mechanism showed a slight reduction in accuracy in some scenarios, it still proved useful in other parameter combinations, such as when a lower learning rate (0.00005) and a batch size adjusted to 64 were used.

Based on the results achieved, several improvements and new directions can be explored in future works to further enhance violence detection in videos and expand the applications of this technology. The experiments conducted focused solely on the use of videos, but it would be interesting to expand to multimodal data, combining audio, image, and even sensor information, which could contribute to a more robust analysis of the scene.

\newpage
\bibliographystyle{sbc}
\bibliography{sbc-template}

\end{document}